\documentclass{amia}
\usepackage[superscript,biblabel]{cite}
\usepackage{graphicx}
\usepackage[labelfont=bf]{caption}
\usepackage{subcaption}
\usepackage{algorithm}
\usepackage{algorithmic}
\usepackage{multirow}
\usepackage{caption}
\usepackage{subcaption}

\floatname{algorithm}{Block}

\begin{document}

\title{TRESTLE: Toolkit for Reproducible Execution of Speech, Text and Language Experiments}

\author{Changye Li, MS\textsuperscript{\rm 1},
    Weizhe, Xu, BS\textsuperscript{\rm 2},
    Trevor Cohen, PhD, FACMI\textsuperscript{\rm 2},
    Martin Michalowski, PhD, FAMIA\textsuperscript{\rm 3},
    Serguei Pakhomov, PhD\textsuperscript{\rm 4}}

\institutes{
    \textsuperscript{\rm 1} Institute of Health Informatics, University of Minnesota, Minneapolis, MN;
    \textsuperscript{\rm 2} Biomedical and Health Informatics, University of Washington, Seattle, Washington;
    \textsuperscript{\rm 3} School of Nursing, University of Minnesota, Minneapolis, MN;
    \textsuperscript{\rm 4} College of Pharmacy, University of Minnesota, Minneapolis, MN;
}

\maketitle

\section*{Abstract}

\textit{The evidence is growing that machine and deep learning methods can learn the subtle differences between the language produced by people with various forms of cognitive impairment such as dementia and cognitively healthy individuals. Valuable public data repositories such as TalkBank have made it possible for researchers in the computational community to join forces and learn from each other to make significant advances in this area. However, due to variability in approaches and data selection strategies used by various researchers, results obtained by different groups have been difficult to compare directly. In this paper, we present TRESTLE (\textbf{T}oolkit for \textbf{R}eproducible \textbf{E}xecution of \textbf{S}peech \textbf{T}ext and \textbf{L}anguage \textbf{E}xperiments), an open source platform that focuses on two datasets from the TalkBank repository with dementia detection as an illustrative domain. Successfully deployed in the hackallenge (Hackathon/Challenge) of the International Workshop on Health Intelligence at AAAI 2022, TRESTLE provides a precise digital blueprint of the data pre-processing and selection strategies that can be reused via TRESTLE by other researchers seeking comparable results with their peers and current state-of-the-art (SOTA) approaches. }

\section*{Introduction}

In the ``Last Words'' letter to ``Computational Linguistics'' in 2008, Pedersen pointed out that the computational linguistics community  was experiencing a reproducibility crisis \cite{pedersen-2008-last}. In that letter, Pedersen provided strong arguments that all published computational linguistic research needs to be accompanied by working software to enable its replication in order to be credible, and that it was ``unreasonable to expect that reproducibility be possible based on the description provided in a publication.'' Ten years later, in 2018, another group of researchers decided to follow up on Pedersen's ``last words'' to investigate the extent to which workers in computational linguistics were willing and able to share their code for the sake of reproducibility. Wieling et al.~\cite{wieling-etal-2018-squib} surveyed 395 publications and found that the code was available either immediately or upon request for only one third of these papers. Furthermore, when they tried to replicate the results for a selection of 10 papers, they were only able to do so for six papers and obtained the exact same results as had been published for only one. These results highlight the magnitude of this persistent problem that is not unique to the computational linguistics community and has been noted in the machine learning (ML) \cite{Kapoor2022}, psychology \cite{Yong2013}, and biomedical natural language processing (NLP) \cite{cohen-etal-2018-three, Digan2020, mieskes:hal-02282794} research fields as well.

The work presented in this paper addresses the broader problem of reproducibility by focusing on a specific subproblem of replicability as set forth by Cohen et al. \cite{cohen-etal-2018-three} in at least one narrowly defined interdisciplinary area of research - computational approaches to characterizing changes in speech and language characteristics caused by cognitive impairment resulting from neurodegenerative conditions such as the Alzheimer's disease (AD). 

This is an important area to address because AD is a debilitating condition with no known cure that affects every aspect of cognition, including language use. Over 50 million people have been diagnosed with AD dementia, and this number is anticipated to triple by 2050 \cite{patterson2018state, prince2016world, world2017global}. Previous studies \cite{lyu2018review, petti2020systematic} have demonstrated that machine learning methods can learn to distinguish between language from healthy controls and dementia patients, automatic analysis of spoken language can potentially provide accurate, easy-to-use, safe, and cost-effective tools for monitoring AD-related cognitive markers. However, a persistent challenge in this work has been the difficulty involved in reproducing prior work and comparing results across studies on account of the use of different diagnosis-related subsets (i.e., probable vs. possible dementia), aggregation strategies (i.e., one vs. multiple transcripts per participant), performance metrics and cross-validation protocols. This challenging issue is exacerbated by the fact that space available for publication of results is typically highly limited and even when a publication venue allows appendices, the description of the methods provided by authors can be highly variable and subject to misinterpretation and uncertainty when trying to reproduce the methods. Consistent with previous finding \cite{wieling-etal-2018-squib}, some researchers provide code while others do not, and the code that is provided typically includes only the implementation of core machine learning methods and does not include scripts needed for data selection and exact execution of validation strategies.

To address this challenge, we developed TRESTLE (\textbf{T}oolkit for \textbf{R}eproducible \textbf{E}xecution of \textbf{S}peech \textbf{T}ext and \textbf{L}anguage \textbf{E}xperiments) for DementiaBank (DB), one of the most popular repositories to host data for the computational linguistics and machine learning communities to build state-of-the-art (SOTA) models on identifying subtle language differences used by dementia patients and healthy controls\footnote{To see a full list of publications that uses data from DementiaBank, see \url{https://dementia.talkbank.org/publications/bib.pdf}}. Particularly, TRESTLE supports data pre-processing for the Pitt corpus \cite{becker1994natural} and other corpora such as transcripts from the Wisconsin Longitudinal Study (WLS) \cite{herd2014cohort} - both are formatted using the CHAT \cite{10.1162/coli.2000.26.4.657} protocol.\footnote{For more details about CHAT protocol, please check the manual here:\url{https://talkbank.org/manuals/CHAT.pdf}}

TRESTLE provides an opportunity for researchers to submit a manifest that includes the precise pre-processing parameters, data selection, and user-defined criteria for ``dementia" and ``control". Therefore, individuals can freely design their own pre-processing parameters and use \textit{exactly the same} data that their peers have provided, allowing for comparable and reproducible evaluation outcomes for their analytical models. 

To the best of our knowledge, this is the first toolkit that provides the infrastructure to enable direct comparisons between experiments conducted on DementiaBank datasets While it is currently designed and tested with data contained in the DementiaBank portion of the TalkBank\cite{macwhinney2007talkbank} repository\footnote{\url{https://www.talkbank.org/}}, it can be easily extended to other public datasets following the CHAT protocol to facilitate reproducibility and comparability of experimental results and the ease of establishing and improving the SOTA in the ML research community. The code for the toolkit is publicly available on GitHub\footnote{\url{https://github.com/LinguisticAnomalies/harmonized-toolkit}}.


\section*{TRESTLE Design Overview}

In theory, if researcher B intends to reproduce the results of methods developed by researcher A, all that researcher B would need to do is ask researcher A for a copy of the data used to obtain the results. In practice, there are many barriers to executing this scenario including the fact that the owners of even publicly available datasets typically do not allow individual researchers to redistribute their data. Therefore, if researcher A makes any modifications to the original data for the purposes of experimentation, these modifications remain with researcher A, as they are not typically propagated back to the original dataset. Researcher B wishing to replicate and improve upon A's results has to obtain the original data from the owner of the data and figure out how to make the same modifications to the original data as were made by researcher A. While researcher A typically does provide in a publication the information describing the data selection and modification decisions, researcher B still has to essentially reconstruct these modifications. Clearly, this situation is error-prone and not conducive to making rapid scientific progress.   

The main motivation for creating TRESTLE stems from the need for a convenient and error-resistant way of communicating the details of a researcher's experimental design to other researchers so they can replicate the experimental conditions in order to test their own methods and compare results to those obtained by previous researchers. Motivated by this need for replicability, the key design feature of TRESTLE is the generation of a machine-readable manifest that captures all of the data selection and pre-processing decisions that were made while running an experiment on the supported datasets. The manifest is intended to be disseminated along with publishing the results of experiments and used as a blueprint to replicate the exact same experimental set up and conditions by others. The objective is to avoid the situation in which a group of researchers develops a new machine learning algorithm for discrimination between dementia cases and controls based on speech and language features, experimentally validates the algorithm on a dataset and publishes the results but another group is not able to reproduce their results because of either insufficient information provided in the publication or misinterpretation of the information or both. An even worse situation may arise where the results are replicated (e.g. same or similar metrics are obtained) but the experimental conditions differ in some subtle ways. Both of these scenarios may lead to meaningless comparisons or significant difficulty in conducting meaningful comparisons and thereby hindering the research community's ability to build on each other's work.     

TRESTLE is also designed to make pre-processing decisions as explicit as possible while providing the flexibility for researchers to add their own pre-processing scripts needed to replicate their results. The motivation for providing this functionality for TRESTLE is secondary to the main motivation for replicating the experimental conditions because pre-processing could be considered a part of one's methodology. For example, including pause fillers (um's and ah's) in training a neural model of speech affected by dementia may be viewed as a novel methodological decision that would contribute to better classification accuracy. As such, pre-processing in general may not lend itself well to standardization. However, in more complex scenarios in which pre-processing itself involves using statistical models or other tools with multiple user-controlled parameters (e.g., target audio sampling rate, noise reduction techniques, etc.) it is also important to capture these parameters precisely and explicitly and provide them together with any software code to subsequent researchers so as to enable them to reproduce these methods.  

\begin{figure}[htbp]
\begin{subfigure}{0.5\textwidth}
  \centering
  \includegraphics[width=\linewidth]{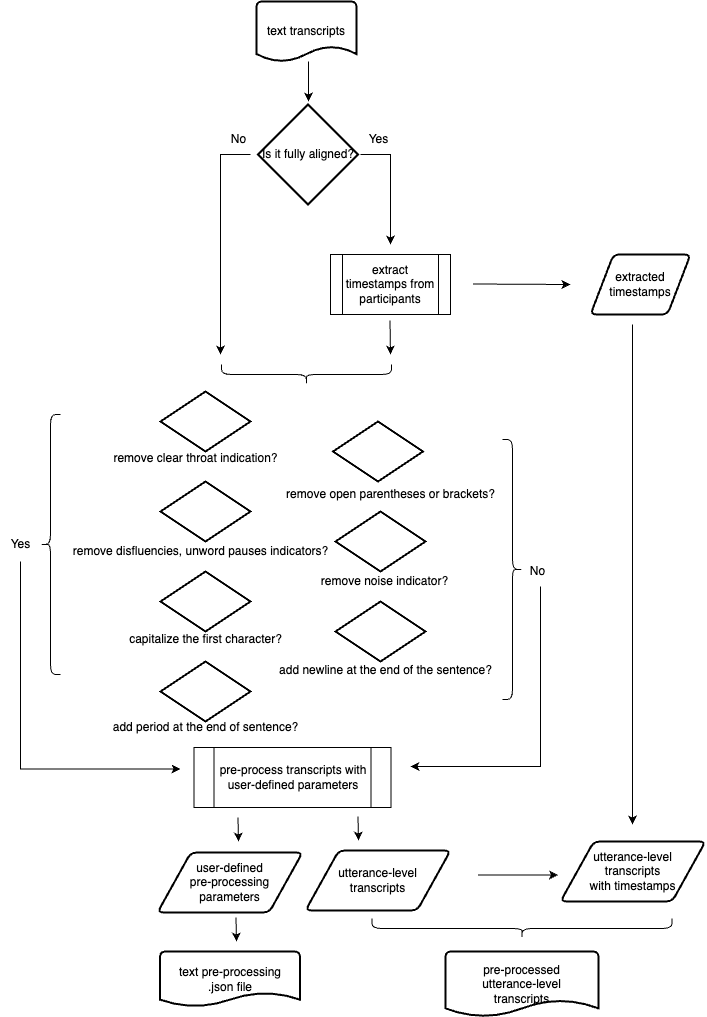}  
  \caption{The overview of the text pre-processing submodule}
  \label{fig:text}
\end{subfigure}
\begin{subfigure}{0.5\textwidth}
  \centering
  \includegraphics[width=\linewidth]{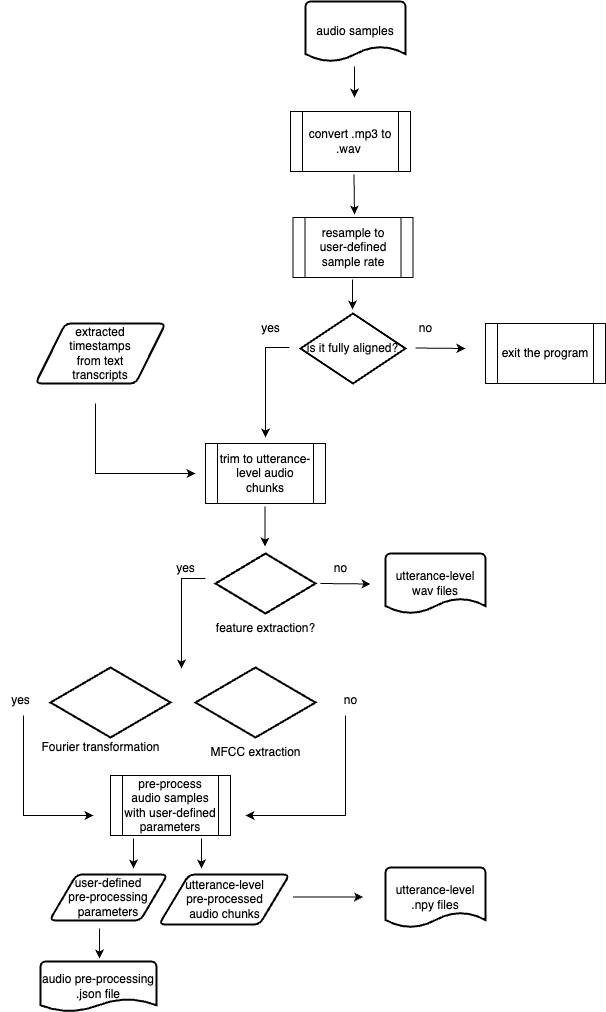}  
  \caption{The overview of the audio pre-processing submodule}
  \label{fig:audio}
\end{subfigure}
\caption{TRESTLE design overview}
\label{fig:trestle_overview}
\end{figure}

The parameters used during sound and text pre-processing/conversion are also stored in the manifest file. In addition to the generation of the manifest, TRESTLE comes with a set of standard pre-processing utilities for text and audio, as demonstrated in Figure~\ref{fig:trestle_overview}. TRESTLE is divided into two sub-modules, a)  pre-processing text data (Figure~\ref{fig:text}), and b) pre-processing audio data (Figure~\ref{fig:audio}) that is fully aligned with the corresponding text transcript. Each sub-module contains a number of parameters that users can define in their own pre-processing manifest. Block~\ref{alg:text_module} and Block~\ref{alg:audio_module} show the general flow of using TRESTLE for pre-processing text and audio samples, respectively.

\begin{algorithm}
\caption{General flow of TRESTLE \textbf{text }pre-processing sub-module. \textit{Italic} indicates inputs from users}
\label{alg:text_module}
Which dataset you are pre-processing? wls or db?: \textit{db}

Where are the .cha files located?: \textit{file-locations}

Remove 'clear throat'? (Y/N): \textit{y}

Remove open parentheses e.g, (be)coming? (Y/N): \textit{y}

Remove open square brackets eg. [: overflowing]? (Y/N): \textit{y}

Remove disfluencies prefixed with '\&'? (Y/N): \textit{y}

Remove unintelligible words? (Y/N): \textit{y}

Remove pauses eg. (.) or (..)? (Y/N): \textit{y}

Remove forward slashes in square brackets? (Y/N): \textit{y}

Remove noise indicators e.g. \&=breath? (Y/N): :\textit{y}

Remove square brackets indicating an error code? (Y/N): \textit{y}

Remove all non-alphanumeric characters? (Y/N): \textit{y}

Replace multiple spaces with a single space? (Y/N): \textit{y}

Capitalize the first character? (Y/N): \textit{y}

Add period at the end of every sentence? (Y/N): \textit{y}

Add newline at the end of every sentence? (Y/N): \textit{n}

You data will be stored as .tsv file. Please enter the output path and file 
name for your pre-processed transcripts: \textit{output.tsv}

Please stand by, your pre-processing script will be generated shortly...

Your text pre-processing json file has been generated!

Running text pre-processing script now...

Your dataset is now pre-processed!
\end{algorithm}

Block~\ref{alg:text_module} demonstrates the pre-processing features currently supported by TRESTLE. As illustrated in Block~\ref{alg:trans}, the raw input CHAT (\texttt{.cha}) file contains several tags indicating participants' behavior during the interview. TRESTLE allows users to remove tags/indicators such as clear throat indicator, open parentheses or brackets, noise, disfluencies, non-words or pauses from the verbatim transcript, if desired. Furthermore, users can choose whether or not to capitalize the first character of each sentence, or add newline at the end of the sentence. Depending on the type of analysis the user intends to do, some or all of these extra-linguistic or para-linguistic elements may need to be either removed or used in the analysis, as demonstrated in several previous studies \cite{orimaye2017predicting, cohen-pakhomov-2020-tale, li-etal-2022-gpt}.

These binary user-controlled parameters are stored in the manifest file in JSON format. Other TRESTLE users can apply the same pre-processing parameters to raw transcripts by using this manifest file, or modify the manifest if comparability to previous work is not desired, giving the authors the option to choose their own criteria but ensure that the criteria are explicit and can be subsequently precisely replicated by others. 

\begin{algorithm}
\caption{General flow of TRESTLE \textbf{audio} pre-processing sub-module. \textit{Italic} indicates inputs from users}
\label{alg:audio_module}
Which dataset you are pre-processing? wls or db?: \textit{db}

Where are the .mp3 files located?: \textit{file-locations}

Where do you want to store the trimmed audio segments? \textit{audio-segments-locations}

Enter sample rate: \textit{16000}

Feature extraction methods, selecting from FTT or MFCC or NONE: \textit{ftt}

Enter number of FTT windows size or MFCC, 0 for NONE: \textit{2}

Scaling MFCC? y/n: \textit{n}

Your audio pre-processing json file has been generated!

Running audio pre-processing script now...

Starting to convert .mp3 to .wav

Finished!

Starting to resample audio to target sample rate...

Finished!

Your dataset is now pre-processed!
\end{algorithm}

Some other barriers to replicability stem from the variability in how raw audio data is pre-processed and prepared for ML. For example, the Pitt corpus audio data is in 16 bit, 16 kHz sampling rate (i.e., 256 kilobits/second bit rate) uncompressed WAVE format whereas the WLS data is in compressed MP3 format encoded at 44.1 kHz sampling rate but 124 kilobits/second bit rate - about half the bit rate of the Pitt corpus. It may be important for studies that use the audio from these datasets to convert the audio to a single specific format needed for analysis and with the understanding of the implications of any such conversion for resulting audio quality. In order to enable these conversions in TRESTLE, we included the Sound eXchange\footnote{\url{http://sox.sourceforge.net/}} library for resampling audio samples. TRESTLE additionally supports feature extraction algorithms such as the Fourier transform (FT) and Mel-frequency cepstral coefficients (MFCC).


These user-controlled parameters are then applied to each text or audio file from the Pitt or WLS datasets. The text sub-module mere all pre-processed utterance-level\texttt{.cha} transcripts to a \texttt{.tsv} file and saves it to the user-specified destination. Furthermore, when pre-processing a dataset in which the text and audio are fully aligned with each other (i.e., Pitt corpus, partial WLS dataset), the text sub-module maintains a list of timestamps indicating the intervals of test administrator speech to the corresponding \texttt{.json} file for further pre-processing in the audio sub-module. The audio sub-module converts audio files to a user-defined format (e.g. single-channel PCM waveform, sampled at 16 kHz). The audio sub-module also generates utterance-level audio segments. When working with the text sub-module together with the corresponding utterance-level transcripts, TRESTLE enables the followup application for automatic speech recognition (ASR) models.

\section*{Datasets}

\begin{figure}[htbp]
    \centering
    \small
    \includegraphics[scale=0.25]{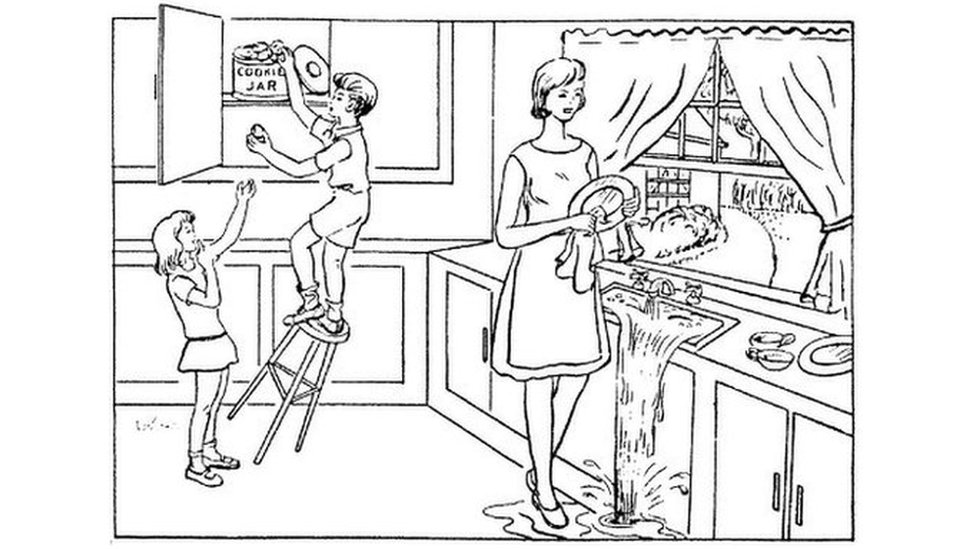}
    \caption{Cookie Theft stimulus}
    \label{fig:cookie_theft}
\end{figure}

TRESTLE currently directly supports data pre-processing for two publicly available and datasets from TalkBank: a) the Pitt corpus\footnote{\url{https://dementia.talkbank.org/access/English/Pitt.html}}, b) and the WLS corpus\footnote{\url{https://dementia.talkbank.org/access/English/WLS.html}}. The details of the two datasets are included in Table~\ref{tab:data}, and a sample transcript formatted in the CHAT protocol from Pitt corpus is shown in Block~\ref{alg:trans}. These datasets, including raw audio, manual transcripts, linguistic annotations, and metadata with demographic, clinical and neuropsychological test characteristis are publicly available from TalkBank.

The Pitt corpus contains audio recordings and manually transcribed transcripts of neuropsychological tests including the ``Cookie Theft'' picture description task from the Boston Diagnostic Aphasia Examination \cite{goodglass1983boston}. In this task, participants were asked to describe everything they see occurring in Figure~\ref{fig:cookie_theft}. Participant responses were audio recorded and subsequently transcribed verbatim. Participants were tested multiple times resulting in multiple transcripts per participant. In total, there are 242 recordings/transcripts from the 99 healthy controls and 257 recordings/transcripts from the 169 participants with AD-related diagnoses. Neurological examination results, including results of the Mini-Mental State Exam (MMSE) and Clinical Dementia Rating (CDR) are also included in the Pitt corpus.

\begin{table}[htbp]
\centering
\small
\caption{Dataset description.}
\begin{tabular}{|ll|l|l|}
\hline
\multicolumn{2}{|l|}{\textbf{Characteristics}}                     & \textbf{Pitt} & \textbf{WLS} \\ \hline
\multicolumn{2}{|l|}{Age, mean (SD)}                     & 69.2 (8.9) & 70.4 (4.4) \\ \hline
\multicolumn{1}{|l|}{\multirow{2}{*}{Gender, \textit{n} (\%)}} & Male & 200 (39.2) & 694 (50.7)\\ \cline{2-4} 
\multicolumn{1}{|l|}{}                  & Female & 310 (60.8) & 675 (49.3) \\ \hline 
\multicolumn{2}{|l|}{Education, mean (SD)}   & 12.5 (3.1)& 13.5 (3.1)\\ \hline
\multicolumn{2}{|l|}{MMSE, mean (SD)}   & 20.7 (7.4) & NA (NA) \\ \hline
\end{tabular}

\label{tab:data}
\end{table}

\begin{algorithm}
\caption{The first few lines of a sample transcript from the Pitt corpus in the CHAT protocol. The morphology and grammar tiers have been omitted for readability. The integers at the end of each line represent start and end times of the utterance in that line. Please check CHAT protocol for more details about tagging in the transcript.}
\label{alg:trans}
@PID:	11312/t-00002420-1\\
@Begin\\
@Languages:	eng\\
@Participants:	PAR Participant, INV Investigator\\
@ID:	eng$\vert$Pitt$\vert$PAR$\vert$57;$\vert$male$\vert$ProbableAD$\vert\vert$Participant$\vert$18$\vert\vert$\\
@ID:	eng$\vert$Pitt$\vert$INV$\vert\vert\vert\vert\vert$Investigator$\vert\vert\vert$\\
@Media:	001-0, audio\\
@Comment:	another audio testing file overlaps in background\\
*INV:	this is the picture . 0\_2581 \newline
*PAR:	mhm . {[+ exc]} 2581\_3426\newline
*INV:	just tell me everything that you see happening in that picture . 3426\_6661\newline
*PAR:	+$<$ alright . {[+ exc]} 6000\_6897\newline
*PAR:	there's \&um a young boy that's getting a cookie jar . 6897\_12218\newline
*PAR:	and it [//] he's \&uh in bad shape because \&uh the thing is fallin(g) over . 12218\_18718\newline
*PAR:	and in the picture the mother is washin(g) dishes and doesn't see it . 18718\_24822\newline
...\newline
@End
\end{algorithm}

The WLS is a longitudinal study of 694 men and 675 women who graduated from Wisconsin high schools in 1957, where the participants were interviewed up to six times between 1957 and 2011. Cognitive evaluations and ``Cookie Theft'' picture description task were introduced to the later rounds of interview on WLS, which are presented in the CHAT-formatted (\texttt{.cha}) files. All of the participants in the WLS were considered to be cognitively healthy upon entry into the study. Some may developed dementia in later years; however, the neurological diagnostic information is not currently publicly available.

Defining the ``dementia'' and ``control'' categories is not entirely straightforward and creates a barrier to reproducibility even if the criteria are described. For example, typical studies involved with the Pitt corpus focus on 169 participants classified as having \textit{possible} or \textit{probable} AD based on clinical or pathological examination, as well as 99 healthy controls. However, 10 of 99 healthy controls later acquired a dementia-related diagnosis - 7 of 10 being diagnosed with probable AD and the remaining 3 having an indeterminate diagnostic status at baseline. This complicates data analysis since individuals' diagnostic statuses may change over time and how this change is treated in a given study may significantly affect the results. The paucity of neurological diagnoses in the WLS also complicates further data analysis. One way to categorize WLS participants into those with potential cognitive impairment and those without is to use the available verbal fluency neuropsychological test scores \cite{10.3389/fcomp.2021.642517}, as verbal fluency (ability to name words belonging to a semantic category) is significantly impaired in dementia and has been recommended for clinical use as a screening instrument for dementia \cite{Canning556}. However, various verbal fluency score cutoffs for dementia have been proposed in the literature and different authors may follow the literature that they trust in selecting the cutoffs. It would not be reasonable to try to impose a single specific cutoff on all studies using the WLS data. 


\section*{Results}

We deployed TRESTLE at the Data Hackallenge\footnote{\url{https://w3phiai2022.w3phi.com/hackathon.html}} of the International Workshop on Health Intelligence\footnote{\url{https://w3phiai2022.w3phi.com/index.html}}, which was co-hosted at AAAI 2022. During the hackallenge, each group of participants used TRESTLE to generate specific subsets of the data along with the pre-processing TRESTLE manifest. Each group was instructed to select data samples from the Pitt or WLS set (or both) using the criteria provided in the corresponding metadata. Each group was also asked to label each selected data sample as ``positive'' (``dementia'') or ``negative'' (``controls'') based on their preferred criteria. Each group was also asked to develop an analytical method (pipeline) of their choosing for discriminating between those categories. Finally, each group was asked to to select an evaluation strategy of their choosing for their analytical method. In the second phase of the hackallenge, each team was asked to evaluate the other group's pre-processing manifests and run their analysis pipeline using the data selection, category definition, and evaluation strategy information provided in the other group's manifest to replicate the other group's experimental design so that the results could be directly compared.

We provided a baseline manifest, following the current SOTA on such tasks with text transcripts \cite{Balagopalan2020ToBO}. For the baseline system, we fine-tuned BERT \cite{Devlin2019BERTPO} on the Alzheimer's Dementia Recognition through Spontaneous Speech (ADReSS) \cite{Luz2020AlzheimersDR} dataset, which is a subset of the Pitt corpus that is matched for age and gender. The baseline manifest with evaluation metrics is shown in Block~\ref{alg:baseline}. Our baseline accuracy and AUC were both 0.77.

\begin{algorithm*}[htbp]
\caption{Sample TRESTLE's baseline manifest in json. Full baseline manifest is available on TRESTLE's GitHub repository}
\label{alg:baseline}
\{

``pre\_process'': ``scripts/text\_process.json'', \hfill \textit{\# pointing to the user-defined text pre-processing parameters}

``data\_uids'':[``001-2'', ``005-2'', ``006-4'', ...], \hfill \textit{\# the sample list of ADReSS dataset, where $-n$ represents the $n$-th visit}

``positive\_uids'': [``001-2", ``005-2", ``010-3", ``018-0", ...], \hfill \textit{\# the sample list of ``dementia'' cases from ADReSS training and test set, where $-n$ represents the $n$-th visit}

``training\_uids": [``001-2", ``005-2", ``006-4", ...], \hfill \textit{\# the sample list of  ADReSS training set, where $-n$ represents the $n$-th visit}

``test\_uids": [``035-1", ``045-0", ``049-1", ...], \hfill \textit{\# the sample list of  ADReSS test set, where $-n$ represents the $n$-th visit}

``method": ``fine-tune BERT", \hfill \textit{\# very short description of method}

``evaluation": \{``ACC": 0.77, ``AUC": 0.77\}\hfill \textit{\# evaluation metrics used for the reported method}

\}
\end{algorithm*}


In addition to our group, two other groups participated in the hackallenge. Both of these teams decided to use both text transcripts and audio recordings from the Pitt corpus and WLS; however, as we anticipated, the two groups had significantly different approaches to selecting data subsets, criteria for classification, and evaluation metrics and strategies. Table~\ref{tab:data-selection} demonstrates the differences between the data selection strategies between two teams who made the final submission. Both team successfully evaluated their methods on manifests of our and the other team and outperformed our baseline model performance with their own models, as seen in Table~\ref{tab:performance}.

\begin{table}[htbp]
\centering
\small
\caption{Criteria defined by the data hackallenge participants for the data selection.}
\begin{tabular}{|l|l|p{11cm}|}
\hline
\textbf{Team}                 & \textbf{Dataset} & \textbf{Cutoff} \\ \hline
Baseline                  & Pitt & ADReSS subset of Pitt corpus \\ \hline
\multirow{2}{*}{Team 1} & Pitt & MMSE $\le$ 24 as dementia, otherwise healthy controls \\ \cline{2-3} 
                 & WLS & Verbal fluency score 16 for individuals aged $<$ 60, 14 for age between 60 and 79, 12 for age $\ge$ 79 as dementia group, otherwise healthy controls \\ \hline
\multirow{2}{*}{Team 2} & Pitt & Diagnosis code 100 as dementia group, diagnosis code 800 as healthy controls \\ \cline{2-3} 
                  & WLS & Category fluency test score of 21 as cutoff \\ \hline
\end{tabular}

\label{tab:data-selection}
\end{table}

\begin{table}[htbp]
\centering
\small
\caption{Best model performances from participants of data hackallenge. Note that these results should be compared in light of the differences in the cutoffs to define categories, as shown in Table~\ref{tab:data-selection}, and differences in the analytical model design. Please refer to the workshop proceeding \cite{workshop} for more details.}
\begin{tabular}{|l|lll|}
\hline
\multirow{2}{*}{\textbf{Team}} & \multicolumn{3}{l|}{\textbf{Performance}}                            \\ \cline{2-4} 
                 & \multicolumn{1}{l|}{Accuracy} & \multicolumn{1}{l|}{AUC} &  F1\\ \hline
Baseline                  & \multicolumn{1}{l|}{0.77} & \multicolumn{1}{l|}{0.77} & NA \\ \hline
Team 1                  & \multicolumn{1}{l|}{0.94} & \multicolumn{1}{l|}{0.92} & 0.84 \\ \hline
Team 2                  & \multicolumn{1}{l|}{0.84} & \multicolumn{1}{l|}{0.92} & 0.77 \\ \hline
\end{tabular}

\label{tab:performance}
\end{table}

\section*{Discussion}
The results of the hackallenge were encouraging as the teams were able to use TRESTLE to achieve directly comparable results to those of the other teams without having to request any additional information or code. One of the key advantages that this hackallenge experiment has demonstrated is the elimination of any uncertainty in comparing results. TRESTLE facilitates the ability to compare results across multiple studies by providing all the necessary context for doing so - including the cutoffs used to define diagnostic categories. If the categories are not defined the same way in two studies, then by definition, the results of these studies cannot be directly compared. TRESTLE provides the information necessary to make this determination unambiguously. The main purpose of the toolkit, however, is to enable researchers to replicate the experimental setup, especially the data selection, exactly as performed by another team so that the only difference between the studies is the classification algorithm. 

TRESTLE presented here has several limitations. First, it only supports data pre-processing for the Pitt corpus and WLS set and does not support pre-processing of the remaining data in DementiaBank. Secondly, the Pitt corpus and the WLS  data are in American English, and many participants of these two studies are representative of White, non-Hispanic American men and women with an average of 12 years education. As result, TRESTLE currently has limited applicability to other ethnic groups and languages, though this may change as data from more diverse samples become available. Thirdly, TRESTLE runs two sub-modules using bash scripts; it may make TRESTLE more difficult to use for researchers who have less experience with programming. Finally, while TRESTLE only supports text or audio sample pre-processing specified on the ``Cookie Theft'' picture description task of the Pitt and WLS dataset in the current iteration, the design of TRESTLE offers the flexibility to generalize to any CHAT-formatted corpus. We believe it can be further iterated and improved for broader data pre-processing of corpora that are hosted on TalkBank for various downstream linguistic or Natural Language Processing (NLP) tasks, including those involving conversational, childhood language, multi-language and clinical datasets. With our access to Dementia Bank, we choose to focus our initial implementation of TRESTLE on the dementia-related corpus. Showing the feasibility of our approach with these data, we plan to further develop TRESTLE to support more datasets and data formats.

\section*{Conclusion}

To address, at least in part, the pervasive challenge of reproducibility, we created TRESTLE, an application that provides researchers working on detection of speech and language characteristics of dementia with a way to replicate each other's experimental setup and explicitly communicate in a machine-readable fashion the parameters used in data pre-processing. TRESTLE was successfully used for the intended purpose in a hackallenge but clearly needs further development to enable wider adoption by the biomedical NLP and computational linguistic communities. Despite the limitations, TRESTLE also has a number of strengths. It provides the researcher with the ability to convey the details of their experiments (e.g. sample selection, category definition, exceptions) in a very transparent and reproducible fashion. This part of TRESTLE is not limited to the Pitt and WLS datasets and can be easily extended to any text and/or audio collection of data. While using the pre-processing modules requires some programming experience and these modules currently support only the Pitt and WLS datasets, they do encapsulate some of the standard practices, tools and methods for text and audio pre-processing. Last but not least, TRESTLE is an open-source package freely available on GitHub to the machine learning and all other communities to use and contribute to.

\section*{Acknowledgement}

This research was supported by grants from the National Institute on Aging (AG069792). 

\bibliographystyle{vancouver}
\bibliography{amia}  

\end{document}